\documentclass{article}

\usepackage{microtype}
\usepackage{graphicx}
\usepackage{subfigure}
\usepackage{booktabs} % for professional tables

\usepackage{hyperref}

\usepackage[accepted]{icml2023}

\usepackage{amsmath}
\usepackage{amssymb}
\usepackage{mathtools}
\usepackage{amsthm}

\usepackage[capitalize,noabbrev]{cleveref}

\usepackage{multirow}
\usepackage[ruled,vlined]{algorithm2e}

\theoremstyle{plain}

\theoremstyle{definition}

\theoremstyle{remark}

\newcommand{\bff}{\mathbf{f}}
\newcommand{\bx}{\mathbf{x}}
\newcommand{\bh}{\mathbf{h}}

\newcommand{\bbR}{\mathbb{R}}
\newcommand{\bH}{\mathbf{H}}
\newcommand{\bU}{\mathbf{U}}
\newcommand{\bZ}{\mathbf{Z}}

\newcommand{\SF}[1]{\textcolor{red}{SF: #1}}
\newcommand{\maz}[1]{\textcolor{red}{Maz: #1}}
\newcommand{\todo}[1]{{\color{red} [{\bf TODO (Neha)}: #1]}}

\iftrue
\renewcommand{\maz}[1]{}
\renewcommand{\SF}[1]{}
\renewcommand{\todo}[1]{}
\fi

\icmltitlerunning{Identifying Interpretable Subspaces in Image Representations}

\begin{document}

\twocolumn[
\icmltitle{Identifying Interpretable Subspaces in Image Representations}



\icmlsetsymbol{equal}{*}

\begin{icmlauthorlist}
\icmlauthor{Neha Kalibhat}{umd}
\icmlauthor{Shweta Bhardwaj}{umd}
\icmlauthor{Bayan Bruss}{cap}
\icmlauthor{Hamed Firooz}{meta}
\icmlauthor{Maziar Sanjabi}{meta}
\icmlauthor{Soheil Feizi}{umd}
\end{icmlauthorlist}

\icmlaffiliation{umd}{University of Maryland, College Park}
\icmlaffiliation{cap}{Center for Machine Learning, CapitalOne}
\icmlaffiliation{meta}{Meta AI}

\icmlcorrespondingauthor{Neha Kalibhat}{\href{mailto:nehamk@umd.edu}{nehamk@umd.edu}}

\icmlkeywords{Machine Learning, ICML}

\vskip 0.3in
]



\printAffiliationsAndNotice{\icmlEqualContribution} % otherwise use the standard text.

\todo{Evaluate transfer of concepts for different models (number of shared top activating images, and/or percentage of feature successfully transferred)}
\todo{User study for higher number of concepts and showing lowly activated images also}
\todo{User study for both relevant and non-relevant (MTurk)}
\todo{SB: Weighted average to ensure high score for relevant concepts}
\todo{SB: Remove bias towards top-k activated images for a given feature. What about abstract concepts in medium activated images? }
\todo{Camera: Plots for new MTurk Study}

\begin{abstract}
We propose Automatic Feature Explanation using Contrasting Concepts (FALCON), an interpretability framework to explain features of image representations. For a target feature, FALCON captions its highly activating cropped images using a large captioning dataset (like LAION-400m) and a pre-trained vision-language model like CLIP. Each word among the captions is scored and ranked leading to a small number of shared, human-understandable concepts that closely describe the target feature. FALCON also applies \textit{contrastive interpretation} using lowly activating (counterfactual) images, to eliminate spurious concepts. Although many existing approaches interpret features independently, we observe in state-of-the-art self-supervised and supervised models, that less than $20\%$ of the representation space can be explained by individual features. We show that features in larger spaces become more interpretable when studied in groups and can be explained with high-order scoring concepts through FALCON. We discuss how extracted concepts can be used to explain and debug failures in downstream tasks. Finally, we present a technique to transfer concepts from one (explainable) representation space to another unseen representation space by learning a simple linear transformation. Code available at \href{https://github.com/NehaKalibhat/falcon-explain}{https://github.com/NehaKalibhat/falcon-explain}.
\end{abstract}

\begin{figure}[h]
    \centering
    \includegraphics[width = 0.49\textwidth]{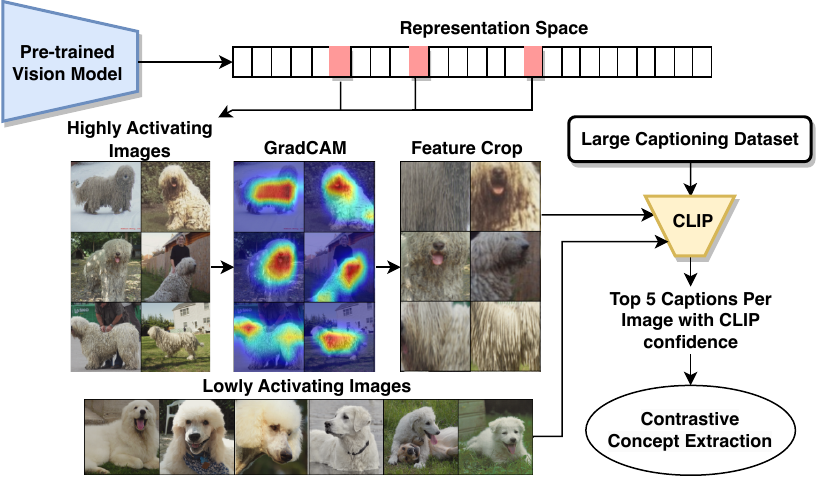}
    \caption{\textbf{Framework of FALCON:} We outline the process of interpreting any given feature(s) in the representation space of a pre-trained model using a probe dataset $\mathcal{D}$ and a captioning dataset $\mathcal{S}$. Taking the set of highly activating images (from $\mathcal{D}$) for the target features we compute their gradient heatmap \cite{gradcam} crops, keeping only the highly activating regions. We compute CLIP \cite{oikarinen2022clipdissect} image representations of the cropped images and text representations of a large captioning dataset (in our case, LAION-400m \cite{schuhmann2021laion400m}). For \textit{contrastive interpretation}, we also caption lowly activating (counterfactual) images. Using cosine similarity, we select the top 5 captions per image and pass them through our concept extraction module (Described in Figure \ref{fig:concept_extract}).
    % \maz{no mention of counterfactual or mined lowly activating images! maybe just mention them in the caption and refer!}
    % \SF{since we also have groups of features to interpret I'd suggest to highight groups in the opening figure}
    }
    \label{fig:framework} 
\end{figure}
\begin{figure*}[t]
    \centering
    \includegraphics[width = \textwidth, trim = {0.2cm, 0.1cm, 0cm, 0.2cm}, clip]{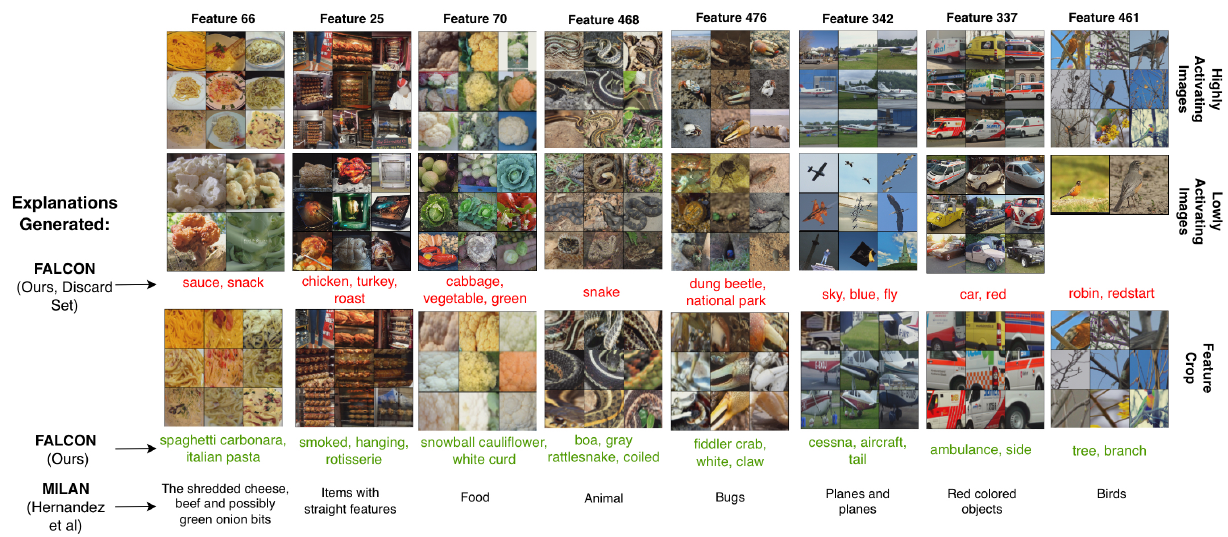}
    \caption{\textbf{Concepts extracted by FALCON for various features in the SimCLR representation space:} We explain various features of the final layer representations of SimCLR \cite{simclr} pre-trained on ImageNet \cite{imagenet} with a ResNet-18 \cite{resnet} backbone (512 features). For each feature, we show the top activating images as well as the lowly activating images. We crop the top activating images to highlight only the activated regions and extract concepts using the approach outlined in Section \ref{sec:feature_interpretation}. The lowly activating images are used to filter spurious concepts using our approach called \textit{contrastive interpretation} (See Equation \ref{eq:lowly}).}
    \label{fig:concepts}
\end{figure*}
\section{Introduction}
Learning generalizable representations has a growing requirement given the considerable cost of pre-training and inference. More importantly, understanding what is encoded in representations is a necessity for deployment, particularly in medical and safety-critical applications \cite{medicalXAI2021}.
% \maz{better with citation}. 
Large pre-trained self-supervised models \cite{dino, simclr, mocov2, simsiam} have shown successful generalization capability with frozen representations, however, their representation spaces are still not fully understood. Prior works attempt to understand neural features through detailed visualization of concepts \cite{olah2020zoom, olah2017feature, gradcam, 2021, ghorbani2019automatic}. Visualization (via saliency) helps discover various attributes that neurons react to, but can be noisy and greatly ambiguous requiring manual inspection to achieve any useful explanation. Natural language explanations can complement saliency heatmaps by providing a small number of conceptual keywords that accurately describe the salient component. Text-based explanations of model features can also enable scalable analysis of model interpretability. We can automatically identify concept frequency and sensitivity, their contribution in downstream tasks and debug failures modes. We note that such analysis is not easily possible using traditional interpretation methods involving saliency (gradient heatmaps). One way to achieve automatic text-explanation is by using supervision datasets \cite{netdissect, hern2022natural} with fine-grained conceptual labels for each sample. Such approaches can prove to be expensive, requiring expert annotations. They also may not be generalizable as explanations can be dataset-specific. 

In the first part of this paper, we propose \textbf{Automatic Feature Explanation using Contrasting Concepts (FALCON)}, a framework to explain neural features, with no densely-labelled dataset or human intervention. We mainly study final-layer self-supervised representations as they contain no label-bias, however, our approach is model-agnostic and can be extended to any deep neural feature. We are also particularly interested in understanding final-layer representations since they alone are accessible to downstream tasks, and their richness and quality is shown to be essential for better generalization \cite{bordes2021high, kalibhat2022measuring, garrido2022rankme}. Nevertheless, our framework is general and can be extended to explain any layer neurons.
% SB: modified previous layer neurons as well. 

\begin{figure*}[h]
    \centering
    \includegraphics[width = \textwidth, trim = {0.2cm, 0.1cm, 0cm, 0.2cm}, clip]{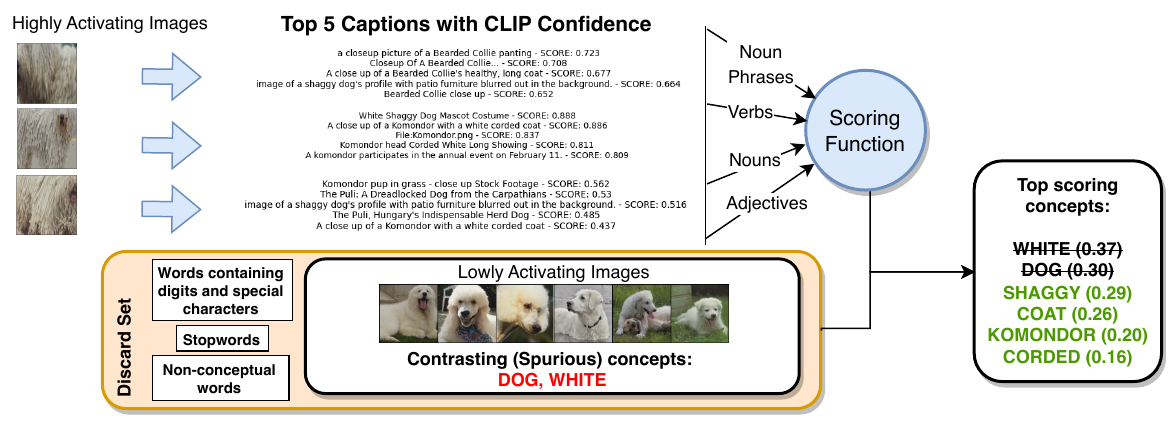}
    \caption{\textbf{Concept extraction in FALCON using contrasting concepts:} We extract a bag of words (nouns, verbs, adjectives) from the top 5 captions (from LAION-400M \cite{schuhmann2021laion400m}) of every image in the set of highly activating images of a given feature. We use a scoring function (Equation \ref{scoring_func}) to extract top scoring words and phrases which we refer to as \textit{concepts}. We also apply \textit{contrastive interpretation} where we discard any concept that is extracted from the lowly activating images (mined through Equation \ref{eq:lowly}). In this case, ``dog" and ``white" are spurious concepts that exist in both highly and lowly activating images, implying that they are not discriminative explanations. Therefore, final set of discriminative concepts include ``shaggy", ``coat", ``komondor" and ``corded" which are all closely related to the given image set.}
    \label{fig:concept_extract}
\end{figure*}

FALCON is described in Figures \ref{fig:framework} and \ref{fig:concept_extract}. For a target feature, we first compute crops of highly activating images from a given dataset (like, ImageNet \cite{imagenet}) based on gradient activation. We then caption each cropped image by matching their CLIP \cite{radford2021learning} image embeddings to the closest CLIP text embeddings from a large captioning dataset (like, LAION-400m \cite{schuhmann2021laion400m}). We collect illustrative captions for each image with high CLIP cosine similarity, without having to train additional captioning models \cite{hern2022natural, 2020, yu2022coca, wiles2022discovering}. The next step in FALCON is described in Figure \ref{fig:concept_extract}, where we show how a compact set of shared, human-understandable \textit{concepts} are extracted from image captions using \textit{Word Score}. We define concepts as the words which closely relate to the attributes that are likely to be encoded by the target feature, based on the set of cropped highly activating images. Unlike prior methods (\cite{oikarinen2022clipdissect}), FALCON is not restricted to output a single concept since features can encode complex physical information which can compose of multiple facets \cite{mu2020compositional}. We recognize, however, that top-ranking concepts can relate to spurious attributes which may not be true descriptors for the target feature, although the attributes exist in most of the highly activating images. Current interpretability techniques \cite{oikarinen2022clipdissect, hern2022natural, netdissect}, tend to produce misguided explanations as they do not account for spuriosity and simply report the highest scoring concept. FALCON eliminates spurious concepts by applying a \textit{contrastive interpretation} technique, where we use lowly activating (counterfactual) images for the target feature whose concepts can be discarded. We therefore produce the minimum sufficient set of concepts that best explain the target. We show the results of successfully annotated features of SimCLR \cite{simclr} in Figure \ref{fig:concepts}. 

\begin{figure*}[h]
    \centering
    \includegraphics[width = \textwidth, trim = {0.1cm, 0.3cm, 0.1cm, 0.3cm}, clip]{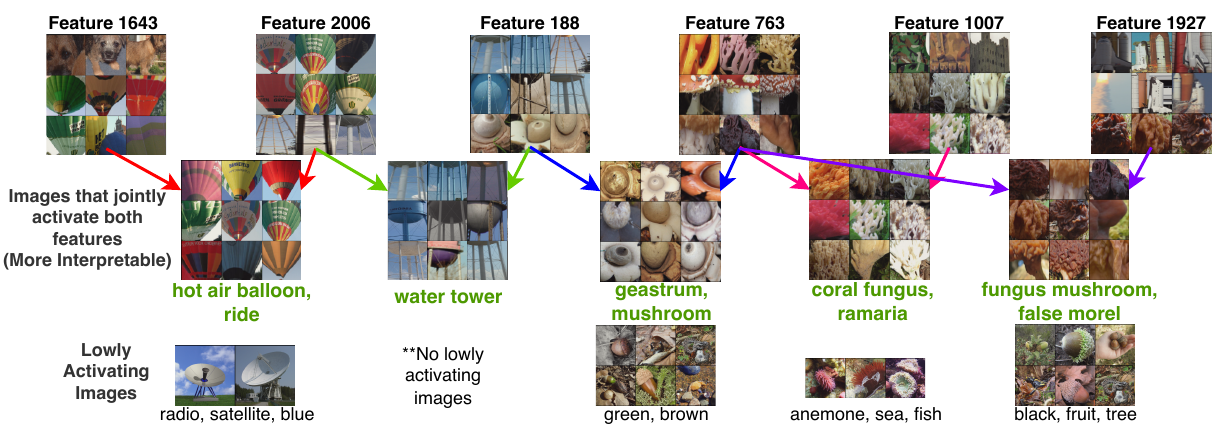}
    \caption{\textbf{Groups of features can be more interpretable than individual features:} In the first panel, we show the highly activating images of some features of DINO \cite{dino} representations trained on ImageNet \cite{imagenet} with a ResNet-50 \cite{resnet} backbone. We observe that the images are highly diverse with seemingly no shared concept, like ``mushrooms" and ``water towers" in feature 188. In the second panel, we observe that images that highly activate pairs of features are significantly more connected. The concepts that our framework extracts are strongly correlated to each group of images. For each feature group, we use the lowly activating images (mined from Equation \ref{eq:lowly}) to filter out spurious concepts.}
    \label{fig:comb_features}
\end{figure*}

In the second part of our paper, we study which features in the representation space can be explained. We observe that individual features that are very strongly activating for an adequate number of samples can correspond to easily detectable concepts. However, such features constitute a very small portion of the whole representation space. We observe that most features activate a diverse set of images where the hidden concept is not apparent (See Figure \ref{fig:comb_features}). We discover that pairs (or groups) of such features are surprisingly more interpretable than individual features. The highly activating images of feature groups are strongly correlated allowing FALCON to produce high scoring concepts. We can therefore explain a much larger portion of the representation space with descriptive and robust concepts.  

We evaluate FALCON through human evaluation on Amazon Mechanical Turk (AMT). We show participants images and their FALCON concepts to collect ground truths (relevant or not relevant) for each concept of each annotated feature. The results from our study show a precision of 0.86 and recall of 0.84 for the top-5 concepts, indicating that FALCON concepts are agreeably explanatory (See Section \ref{sec:evaluation}).
%\maz{refer to section! Are there any baselines here to discuss?}.

Since the extracted concepts are unique physical attributes for only the portions that a given feature encodes, we can decompose the content of any given image into a set of concepts corresponding to different elements (See Figure \ref{fig:image_decomposition}). This helps us understand which physical components of the image have been encoded in its representation. This is also not possible with approaches that conceptualize entire images (like \cite{oikarinen2022clipdissect}). We further utilize concepts to explain failures, like mis-classifications in downstream tasks (See Figure \ref{fig:failure_mode}). By discovering the most contributing concepts in classification, we can detect what the model pays attention to while making its prediction and communicate these in terms of human-understandable concepts. This can help practitioners find and debug issues like hard examples, multi-object scenarios and mis-labelled examples. 
% \maz{if this is an important piece do we want to allocate more space?}

Finally, we propose an approach to transfer concepts from an explained representation space to a new representation space by learning a simple, linear transformation. We train a linear head that maps representations from a target (unseen) model to the source (interpretable) model. This function lets us map any interpretable feature (or group of features) in the source model to the corresponding feature (or group of features) in the target model, and transfer the extracted concepts. We show that the top activating images of the features in the new representation space, exactly match the transferred concepts from the source representation space (See Figure \ref{fig:transfer}).

We summarize our contributions below:
\begin{itemize}
    \item We propose Automatic Feature Explanation using Contrasting Concepts (FALCON), an interpretability framework that automatically detects concepts encoded by any feature of image representations, without any labelled datasets or human intervention. 
    \item We show that representation spaces can be largely explained by interpretable feature groups rather than independent features.
    \item We show that concepts can be used to explain failures in downstream tasks and can be transferred across representation spaces with a simple linear transformation.
\end{itemize}

\section{Automatic Feature Explanation using Contrasting Concepts (FALCON)}
\label{sec:feature_interpretation}
\subsection{Image Captioning Using CLIP}
We discuss the general workflow of FALCON to explain features of vision model representations. Let us consider a pre-trained backbone denoted by $\bff_\theta(\cdot)$. For a given input image $\bx$, this model outputs a representation vector of size $r$, i.e, $\bff_\theta(\bx) = \bh \in \bbR^{r}$. Any downstream task only utilizes these representation vectors, therefore, our objective is to provide human-understandable explanations for these features. 

In order to explain features (different indices in $\bh$), we utilize two datasets ; 1) A probing dataset consisting of a diverse set of images ($\mathcal{D}$), and 2) A large text dataset to extract concepts ($\mathcal{S}$). In our experiments, we use ImageNet-1K \cite{imagenet} validation set for $\mathcal{D}$ and LAION-400m \cite{schuhmann2021laion400m} for $\mathcal{S}$, however, the framework of FALCON is general and can be used with other datasets as well. 
% \maz{are the models also trained on imagenet? do we do this on the imagenet test data? this might make it a bit questionable from reviewers.} 

Let us consider the task of explaining the $i^{th} (0 \le i \le r)$ feature in the representation space of a pre-trained vision model $f_\theta(\cdot)$. From the probing dataset, $\mathcal{D}$ of size $N$, we first extract the set of highly activating images for feature $i$ defined by, $\mathcal{T}_i = \{j: h_{ji} > \alpha, 1 \le j \le N \}$, where $\alpha$ is a threshold we empirically select (more discussed in Section \ref{sec:which_features}). As shown in Figure \ref{fig:framework}, for SimCLR \cite{simclr} with a ResNet-18 \cite{resnet} backbone, the set of highly activating images for feature $10$ are images of a certain breed of dogs. We next compute the gradient of feature $i$ with respect to these images using GradCAM \cite{gradcam} as shown. We crop the images keeping only the maximally activating portions by thresholding the GradCAM mask. This set of cropped images as well as a large scale text dataset ($\mathcal{S}$) like LAION-400m, serve as the input to a pre-trained vision-language model, i.e., CLIP (ViT-B/32) \cite{radford2021learning}. LAION-400m is a large, diverse image captioning dataset which has been used to pre-train vision-language models like CLIP.

We define the CLIP text encoder as $g_{tx}(\cdot)$ and image encoder as $g_{im}(\cdot)$. Given our captioning dataset ($\mathcal{S}$) of size $M$, we extract the text embedding matrix denoted by $A \in \bbR^{M \times k}$ where $k$ is the size of the CLIP text embedding space. Since our captioning dataset is fixed for interpreting any feature, we only need to compute its embeddings once. In fact, LAION also provides pre-computed text embeddings on CLIP which saves compute time significantly. We next compute the image embeddings of the cropped highly activating images of feature $i$ denoted by $B \in \bbR^{|\mathcal{T}_i| \times k}$. Using $A$ and $B$, we compute the CLIP confidence matrix, which is essentially the cosine similarity matrix, denoted by $C = BA^T \in \bbR^{|\mathcal{T}_i| \times M}$. Note that both text and image embeddings are L2-normalized before computing $C$. Using $C$, we extract the top 5 captions for each image in $\mathcal{T}_i$. 

\subsection{Contrastive Concept Extraction}
The second component of FALCON involves extracting concepts out of the captioned batch of highly activating images for the given feature. In Figure \ref{fig:concept_extract}, we show the top-5 concepts for three highly activating images along with the CLIP confidence. From each caption, we extract the noun phrases, nouns, verbs and adjectives to form a bag of words. Verbs and adjectives are extracted to qualify complex concepts which cannot be described with nouns alone. We remove all stop words and words containing digits or special characters from the bag. We also prepare a discard word set including general, non-conceptual words like ``photo", ``picture", ``background" etc. Given a word $w$, the word confidence for the $p^{th}$ caption in the $q^{th}$ image is given by, $C_{q,p}^w$ if the word exists in the caption, otherwise $0$. We get the maximum value of $C_{q,p}^w$ for each image ($q$). The \textit{Word Score} is defined as, 
\begin{align}
    \textit{Word Score}^w = \frac{1}{|\mathcal{T}_i|}\sum_{q = 1}^{|\mathcal{T}_i|} \max_p C_{q,p}^w
    \label{scoring_func}
\end{align}
% \maz{can we simplify the notation and explanation? what is $T_i$ and what is the max over?} 
Word Score gives a normalized score for every word among the captions we extract. The best shared concepts describing a given feature $i$ are the highest ranking words, by applying a threshold (in practice, $0.08$). 

\textbf{Contrastive Interpretation:} In practice, the above method of concept extraction provides a number of high-scoring keywords, shared between the highly activating images. However, in many cases, these keywords can be too general or related to high-level spurious attributes which may be common to all the activating images but not necessarily relevant to the feature we want to interpret. Many existing techniques \cite{oikarinen2022clipdissect, netdissect, hern2022natural, mu2020compositional}, do not account for such cases and they only report a single best scoring concept. In FALCON, we overcome this issue by discovering images in $\mathcal{D}$ that share all other concepts with the highly activating images of feature $i$, except the actual concepts that feature $i$ encodes. We refer to these images as \textit{lowly activating counterfactual images}. The concepts extracted out of lowly activating images can be regarded as spurious concepts for feature $i$ and added to the discard set. 

%% (Discuss with Neha) Shweta: notations are not very clear at first glance. 
Let us define the set of feature indices without the index $i$ as $\mathcal{V}_i = \{j: 0 \le j \le r, j \ne i\}$. The mean representation of the highly activating images ignoring the $i^{th}$ feature can be written as $\bh^{\mu} = \textit{mean}_{\mathcal{T}_i}(\bh_{\mathcal{T}_i, \mathcal{V}_i}) \in \bbR^{|\mathcal{V}_i|}$. The set of lowly activating images for the target feature $i$ is given by, 
\begin{align}
\mathcal{L}_i = \{j: h_{ji} < \epsilon, \bh_{j, \mathcal{V}_i} \boldsymbol{\cdot} \bh^{\mu} \ge \beta, 0 \le j \le N \}
\label{eq:lowly}
\end{align}
% \maz{I follow the explanation but not the equation. Simplify and expand notation. Can be moved to appendix if needed!} 
where $\beta$ and $\epsilon$ are limits we select empirically. In our experiments, $\epsilon$ is the mean value of that feature across the population of normalized representations. Since $\beta$ is used to threshold the dot product of representations (excluding the target feature), a larger value for $\beta$ would give us true counterfactuals. We therefore select $\beta$ to be $0.7$. This method of conditional selection gives us lowly activating images that contain all the concepts in the highly activating image set, except the concept represented by the $i^{th}$ feature. We apply FALCON (without feature cropping) to extract concepts out of the lowly activating image set. As shown in Figure \ref{fig:concept_extract}, concepts like ``dog" and ``white" are in lowly activating images. These keywords can be relevant to the highly activating image batch as well, however, they are not discriminative explanations for that feature. Therefore, we include the concepts of lowly activating images in the discard set and arrive at the final minimum sufficient set of concepts ``shaggy", ``coat", ``komondor" and ``corded".

In Figure \ref{fig:concepts}, we show the extracted concepts from FALCON for 8 different features of SimCLR on a ResNet-18 backbone. In cases like Feature $337$, the lowly activating images match almost all the object properties i.e, vehicle or van. However, after extracting concepts, it becomes clear that the feature concept is the side view of an emergency vehicle which is explained by - ``ambulance" and ``side". Contrastive interpretation therefore lets us ignore generic and spurious concepts to derive a compact set of discriminative explanations. 
% \maz{This part of the paragraph is too repretetive. Maybe combine with the previous paragraph.} 
We also compare FALCON with MILAN \cite{hern2022natural}, a recent approach that trains a generative model on a human-annotated fine-grained image region-caption dataset, and uses this model to generate natural language explanations. We observe that FALCON produces more feature-specific concepts compared to the generic high-level explanations of MILAN. We show more annotated features (including supervised and previous-layer features) comparing with MILAN in the Appendix (See Figures \ref{fig:concepts_unseen_data}, \ref{fig:resnet50}). We also discuss the generalizability of concepts to an unseen dataset like STL-10 \cite{stl10} (See Figure \ref{fig:concepts_unseen_data}).

% \maz{We can isntead expand the comparison.} \maz{Also for the plots how much cherry picking di we do? Can we mention something about the cherry picking and also have more stuff in the appendix to prove that this is not cherry picked?}

\section{Which Features are Explainable?}
\label{sec:which_features}
So far we discussed our method to explain individual features, given the representation space of a pre-trained model. In this section, we understand which features in the representation space can be considered as \textit{explainable}. Let us go back to the set of highly activating images for a given feature $i$, defined by $\mathcal{T}_i = \{j: h_{ji} > \alpha, 1 \le j \le N \}$. Note that the representations are all L2-normalized. In order to extract meaningful and generalizable shared concepts with high Word Scores, we require a sufficient number of highly activating images. In our experiments, we select the features where $|\mathcal{T}_i| > 10$. If $\alpha$ is large enough, we may expect the set of highly activating images to be more connected where the feature concept is clearly detectable (See features in Figure \ref{fig:concepts}). 

We choose features with a strong value for $\alpha$ according to the distribution of the representation space of the selected model. The features where $|\mathcal{T}_i| > 10$, only comprises of roughly $20\%$ of the representation space. See Table \ref{tab:count-features} for this percentage for various pre-trained models. Upon empirical inspection of the activated images, we find that thresholding $\alpha$ alone, may not guarantee explainability. Some of the features can correspond to human recognizable concepts (activating correlated images), like the examples shown in Figure \ref{fig:concepts}. While other features, although strongly activated for a sufficient number of samples, correspond to very high level, abstract concepts that are not apparent to humans. We show examples of such features in the top panel of Figure \ref{fig:comb_features}, on DINO \cite{dino} with a ResNet-50 backbone. Although these features are activating with high $\alpha$, the images are quite diverse, making it almost impossible to decipher any shared properties. One possible way to understand such features could be by explaining previous layer neurons in the network which may perhaps encode higher level properties \cite{oikarinen2022clipdissect, mu2020compositional, hern2022natural, netdissect}. This is however computationally inefficient as previous layer features may still activate dissimilar image sets or may correspond to entirely different concepts.

In the second panel of Figure \ref{fig:comb_features}, we make a key observation; images that jointly activate a given pair of features are significantly more related and explainable than those of individual features. For example, visually, we cannot identify any shared property between rockets and morel mushrooms in feature 1927 and similarly, fly argaric mushrooms and underwater coral plants in feature 763. However, when both feature 763 and 1927 are highly activated, the shared concepts become more apparent, showing only morel mushroom textures. When the same feature 763 is jointly activating with another feature like 1007, it corresponds to a totally different concept of coral reef patterns. A similar observation has been made in \cite{elhage2022toy, net2vec}. Note that the threshold for $\alpha$ is the same for both individual and groups of features (for fair comparison in Figure \ref{fig:comb_features}), however, less rigorous $\alpha$ can still be used for groups of features. By observing highly activating images for a combination of features, we can explain a larger portion of the representation space (even by relaxing $\alpha$) compared to independent features. 

\textbf{Automatically discovering all interpretable feature groups:} Given a model $f_\theta(.)$ and a probe dataset $\mathcal{D}$ of $N$ samples, we compute the top activating set of features (group) for every sample (using $\alpha$ as the threshold). We save each feature group and the indices of the samples that highly activate that group. We use the average CLIP cosine similarity of the samples within each group to decide if a group is interpretable or not (using a threshold, $\gamma$). A higher value for average similarity implies that the top activating samples are \textit{similar} with interpretable shared concepts. Other metrics LPIPS \cite{zhang2018unreasonable} can also be used. In Algorithm \ref{alg:int_feat_selection}, we provide PyTorch-like code highlighting the steps required for identifying all possible interpretable feature groups in the representation space of a given model. 

FALCON can be used to extract concepts out of groups of features in the same manner as individual features, with some modifications. First, the feature crop is calculated by taking the intersection of the gradient heat map of each feature individually as shown in Figure \ref{fig:comb_features}. Second, the lowly activating images are mined such that all the features in the group show low activation and the remaining features are close to that of the highly activating representations. That is, Equation \ref{eq:lowly} is updated to compute $\mathcal{L}_{\mathcal{I}}$ where $\mathcal{I}$ represents the feature group. As shown in Figure \ref{fig:comb_features}, FALCON uses the lowly activating images to help in finding discriminative concepts for groups of features that best explain the highly activating images.

In Appendix Section \ref{sec:global_analysis}, we analyze the extracted concepts across various models (supervised and self-supervised) and discuss some key insights. 
\definecolor{commentcolor}{RGB}{110,154,155}   % define comment color
\newcommand{\PyComment}[1]{\fontsize 80\ttfamily\textcolor{commentcolor}{\textbf{\# #1}}}  % add a "#" before the input text "#1"
\newcommand{\PyCode}[1]{\fontsize 80\ttfamily\textcolor{black}{\textbf{#1}}} % \ttfamily is the code font

\begin{algorithm}[h]
\caption{Pytorch-like pseudocode for discovering interpretable feature groups in a given representation space} \label{alg:int_feat_selection}
\SetAlgoLined
    \textbf{Input:} $\bH$ is the set of representations (of the given model $f_\theta(.)$) of $N$ samples in the probing dataset $\mathcal{D}$, \\
    $\alpha$ is a threshold for feature activation, \\
    $\gamma$ is a threshold for interpretable feature groups. \\
    % \fontsize 100
    \PyComment{Identify all feature groups} \\
    \PyCode{groups = \{\}} \\
    \PyCode{for j in range(N):} \\
    \Indp
        \PyCode{group = torch.where(h[j] > alpha)} \\
        \PyCode{groups[group].append(j)} \PyComment{groups[group] is a list} \\
    \Indm
    \PyComment{Filter out interpretable groups} \\
    \PyCode{int\_groups = \{\}} \\
    \PyCode{for group in groups:} \\
    \Indp
        \PyCode{if len(groups[group]) > 10:} \\
            \Indp
                \PyComment{top activating samples for group} \\                
                \PyCode{top\_act\_idx = groups[group]} \\
                \PyCode{clip\_feat = get\_clip\_feat(top\_act\_idx)} \\
                \PyCode{avg\_cos = torch.matmul(clip\_feat, clip\_feat.T).mean()} \\
                \PyCode{if avg\_cos > gamma:} \\
                \Indp
                    \PyCode{int\_groups[group] = groups[group]} \\
                \Indm
            \Indm
    \Indm
    \PyCode{return int\_groups}
\end{algorithm}

\section{Evaluating Extracted Concepts}
\label{sec:evaluation}
FALCON produces a simple, compact set of concepts to describe any explainable feature in an automatic fashion without any human intervention, or densely-labelled datasets. We performed a human study on Amazon Mechanical Turk (AMT) to evaluate the concepts generated by FALCON and provide some quantitative metrics. In each task, we showed the AMT participant the set of highly activating cropped image set (Group A) and the lowly activating image set (Group B) for a target feature and, the top $6$ concepts ranked by FALCON. We asked the participant to - identify the concepts that are related to Group A and not Group B. This lets us assign binary ground-truth labels to each concept as 0 (not related) if it has been chosen by at least 65\% of the participants and 1 (related) otherwise. We can partition the six FALCON concepts for each feature based on their rank such that the first $K$ concepts where $1 \le K < 6$ can be predicted as 1 (related), otherwise 0 (not related). In Table \ref{tab:pr_amt}, we plot the Precision and Recall for each $K$. Precision in our case measures how many of the ``related" concepts predicted by FALCON are actually related according to our human study. Recall measures how many ``related" concepts was FALCON able to predict among the total number of related concepts (from our human study). We observe that the Recall improves from the $4^{th}$ caption, meaning that, the participants agree that the first 4-5 concepts are related to the given set of images. $84.23\%$ (precision at top-6) of all FALCON concepts are considered relevant by our participants. This study confirms that the top ranking concepts generated by FALCON are considered relevant and explainable among humans. We collect ground truths for $600$ concepts each from $3$ participants. We measure the agreement between participants for each feature by averaging the \% overlap of the concepts selected by each participant. The average agreement among the participants is $79\%$. More details about our human study can be found in Appendix Section \ref{sec:mturkstudy}.

Existing methods (MILAN \cite{hern2022natural}, NetDissect \cite{netdissect}) use human-annotated datasets for natural language descriptions. Through FALCON, we automatically extract a minimal sufficient set of noise-free concepts with no human intervention. We performed another user-study on MTurk to provide a quantitative comparison of FALCON with MILAN and Net-Dissect. We display the highly (Group A) and lowly (Group B) activating images for each target feature and requested the participants to select the concept set which best describes the images in Group A but not Group B. We tabulate the percentage of times the concept set of each framework was selected as the best explanation in Table \ref{tab:falcon_vs_baselines}. FALCON performs significantly better than the baselines in our study of $115$ features.
\begin{table}[h]
    \centering
    \caption{\textbf{Precision and Recall for human evaluation of top $K$ concepts:} Using Amazon Mechanical Turk (AMT), we ask human participants to choose the un-related captions, among 6 top-ranking captions for each feature. We use the annotations as ground truth labels (relevant or not relevant) and compare them to the predictions of FALCON at different levels of $K$ (number of predicted concepts labelled as relevant).}
    \resizebox{0.3\textwidth}{!}{
    \begin{tabular}{c|c|c}
    \toprule
        \textbf{Top $K$} &  \multirow{2}{*}{\textbf{Precision (\%)}} & \multirow{2}{*}{\textbf{Recall (\%)}} \\
        \textbf{Concepts} & & \\
        \midrule
        \midrule
         1 & 94.62 & 18.72 \\
         2 & 92.47 & 36.60 \\
         3 & 88.89 & 52.77 \\
         4 & 86.82 & 68.72 \\
         5 & 85.60 & 84.68 \\
         6 & 84.23 & 100.00 \\
    \bottomrule
    \end{tabular}
    }
    \label{tab:pr_amt}
\end{table}
\begin{table}[h]
    \centering
    \caption{\textbf{Comparing explanations generated by FALCON with existing baselines:} We request participants to select the best explanation generated by the following 3 frameworks for a given set of highly and lowly activating images. FALCON beats other baselines by a significant amount.}
    \label{tab:falcon_vs_baselines}
    \resizebox{0.35\textwidth}{!}{
    \begin{tabular}{c|c}
    \toprule
         \multirow{2}{*}{\textbf{Framework}} & \textbf{\% of times selected} \\
         & \textbf{as best explanation} \\
         \midrule
         \midrule
         FALCON & \textbf{86.40} \\ 
         MILAN \cite{hern2022natural} & 13.47\\ 
         Net-Dissect \cite{netdissect} & 0.12\\ 
         \bottomrule
    \end{tabular}
    }
\end{table}

\begin{figure}[h]
    \centering
    \includegraphics[width = 0.49\textwidth, trim = {0.3cm, 0.1cm, 0cm, 0.2cm}, clip]{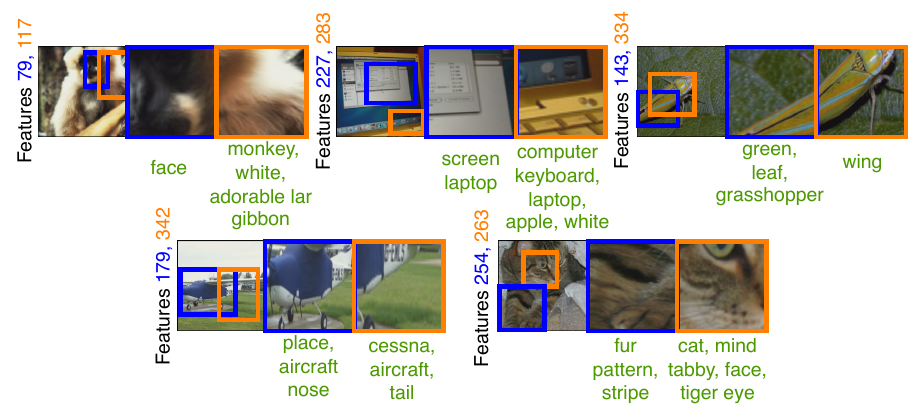}
    \caption{\textbf{Decomposing images into various concepts:} We show some images which highly activate multiple interpretable features. FALCON extracts concepts from feature crops rather than entire images, therefore, each image can be broken down into components, each describing a different physical attribute.  }
    \label{fig:image_decomposition}
\end{figure}  
\begin{figure}[h]
    \centering
    \includegraphics[width = 0.47\textwidth, trim = {0cm, 0.2cm, 0cm, 0.4cm}, clip]{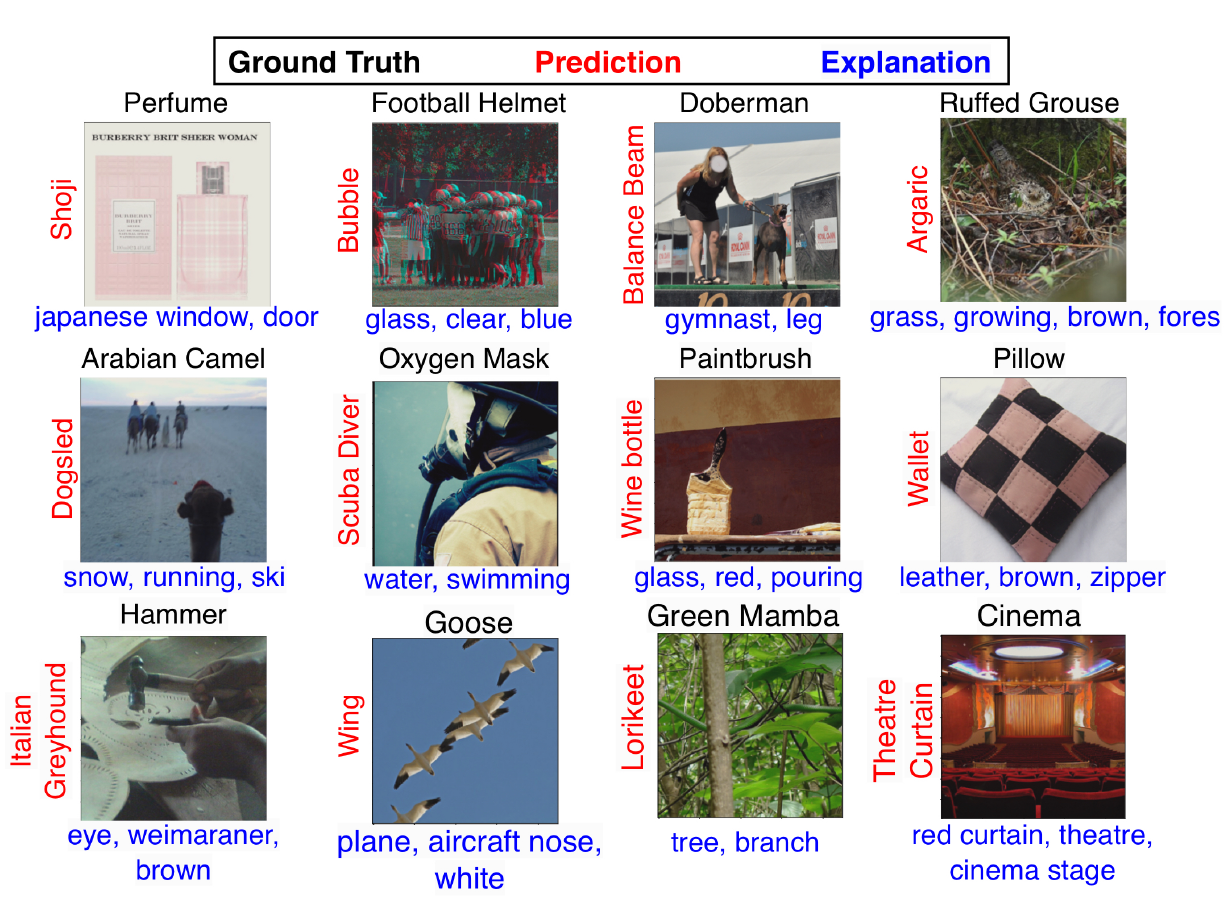}
    \caption{\textbf{Explaining failures in downstream tasks using concepts:} Given SimCLR 
    \cite{simclr} pre-trained on ImageNet \cite{imagenet}, we show some mis-classified examples along with the most contributing concepts for their prediction. This allows us to detect and explain concepts which contributed to a model's decision and help us debug model failures. 
    % \SF{a bit hard to read captions below images; perhaps add "concepts:" and increase spacing between rows; same comment on some other figures } \todo{explain this better with highest and second highest scoring captions}    
    }
    \label{fig:failure_mode}
\end{figure}
\section{Explaining Failure Modes in Vision Models}
An interpretable representation space of a given model, allows us to decompose and label different groups of concepts in any given image. In the previous sections, we found that each interpretable feature (or group of features) encodes only a portion of images that correspond to a unique concept set. Therefore, images with multiple highly activating features \cite{kalibhat2022measuring}, can be decomposed into multiple components, each representing a unique concept. We illustrate this in Figure \ref{fig:image_decomposition}, where we show the feature crop of highly activating features (of SimCLR with ResNet-18) in each image, and their corresponding physical concepts that our framework has extracted. This is only possible because FALCON uses feature crops to discover concepts rather than whole images (unlike CLIP-Dissect \cite{oikarinen2022clipdissect}). 

Another advantage of feature specific concepts is the ability to explain failures in downstream tasks. It is often not obvious what led to a model's prediction without some qualifying explanations. Images in the real-world could contain several spurious attributes interfering with the main content of the image. In such cases, it can be difficult for even experts to localize the exact reason for mis-classification. Moreover, it is often too tedious to have humans make guesses as to what could be the reason for failures as each human can interpret images in a unique manner. With our automatic explanation framework, FALCON, we eliminate this need for human-in-the-loop and can inspect grounded explanations directly.

We consider the task of classification using a linear head defined by the weight matrix $\bU \in \bbR^{o \times r}$, where $o$ is the number of classes. The most contributing features (and corresponding concepts) for a sample $\bx_j$ with prediction $y_j$, can be given by, $\text{arg} \max(\bh_j \odot \bU_{y_j})$. In Figure \ref{fig:failure_mode}, we show some mis-classified examples of SimCLR trained on ImageNet and the most contributing concepts for each prediction. The concepts we find add novel insight into model behavior apart from the readily available information i.e., the image, label and prediction. They help describe the attributes to which model paid attention while making its prediction, potentially helping us automatically debug models at inference time. 

For example, the eighth ``Goose" image, looked more like an aircraft to the model, leading to the prediction ``Wing". This is an example where the model may be spuriously associating shape (like an aircraft) and background information (like the sky) in making its prediction, failing to identify the subtle features of geese. The texture in the ``Perfume" and Football Helmet" images is also an example of spurious attributes. The sixth ``Green Mamba" image can be regarded as a \textit{hard example}, where the core object is largely hidden, causing the model to focus more on concepts like tree and branch. Explanations can also help uncover images which may have multiple ground truths like the eleventh example of ``Cinema" and ``Theatre Curtain" (similar to the images in Figure \ref{fig:image_decomposition}). The ``Pillow" and ``Hammer" images indicate that the training paradigm of the model ignored global object information and made decisions based on local attributes. One possible approach to improve such models relying on spurious correlations is by fine-tuning on synthetic images generated using the relevant FALCON concepts via methods like Stable Diffusion \cite{rombach2021highresolution}. Explanations can also help define optimal training augmentations that could prevent spurious dependencies.

\section{Transferring Concepts to New Representation Spaces}
\label{sec:transfer}
So far, we have discussed the process of feature captioning and concept extraction for a given vision model. We hypothesize that the representations learned by different models can be mapped from one to another. This would allow us to map the features of an explainable representation space to any unseen representation space, without having to re-run our explanation framework. Let us consider the representations of a model that we have extracted concepts for, denoted by $\bH_{source} \in \bbR^{N \times r}$. The representation space of an unseen model can be denoted by $\bH_{target} \in \bbR^{N \times r}$. Using a linear head, our goal is to learn a transformation matrix $\bZ \in \bbR^{r \times r}$, that transforms $\bH_{target}$ to $\bH_{source}$, by solving the optimization,
\begin{align}
    \min_{\bZ} \|\bZ^T \bH_{target} - \bH_{source}\|_2
    \label{eq:transfer}
\end{align}
We solve optimization by training a linear head for only 10 epochs with a learning rate of $1$, using an SGD optimizer. Once the mapping is learned, we can take any explainable feature $i$ in $\bH_{source}$, and find the features in $\bH_{target}$ with have the highest weights in $\bZ$. Hence, the concepts described by feature $i$ in $\bH_{source}$ can be mapped to features in $\bH_{target}$ efficiently. 

\begin{figure}[h]
    \centering
    \includegraphics[width = 0.49\textwidth, trim = {0cm, 0cm, 0cm, 0.3cm}, clip]{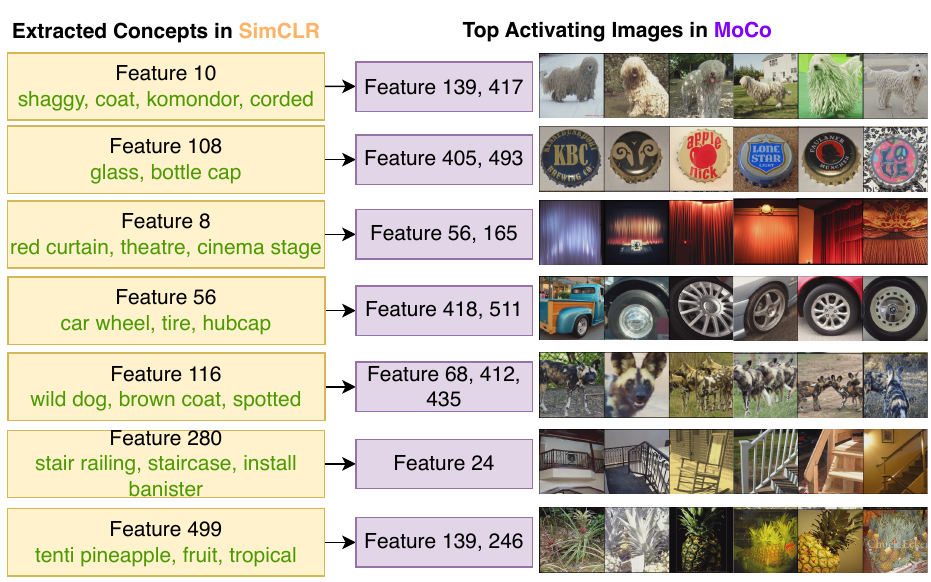}
    \caption{\textbf{Transferring concepts from an explained representation space to an unseen representation space:} We show that representations of self-supervised models can be mapped from one to another by learning a transformation $\bZ$ (See Equation \ref{eq:transfer}). We transfer extracted concepts from SimCLR (source) \cite{simclr}, to an unseen model, MoCo (target) \cite{mocov2} by mapping the source features to the target features with the highest weights in $\bZ$. We observe that the top activating images of the mapped features in the MoCo very closely match the concepts extracted in SimCLR. }
    \label{fig:transfer}
\end{figure}

We confirm that this transformation works by matching the concepts of $\bH_{target}$ to the highly activating images of $\bH_{target}$. As shown in Figure \ref{fig:transfer}, we successfully map individual interpretable features in SimCLR to features in MoCo \cite{mocov2} which is an unseen representation space. The highly activating images in MoCo interestingly contain all of the concepts of the source feature. Note that, features across representation spaces need not have a 1:1 relationship. Similarly, concepts can also correspond to compositional features (as described in Section \ref{sec:which_features}). We do not constrain the sparsity of $\bZ$. In practice, $\bZ$ is not sparse however, it can be considered as \textit{nearly sparse} where most weights are close to zero. When we discover feature maps, we only extract the weights in $\bZ$ if they are large enough ($> \text{mean} + 4 \times \text{std}$ based on the weight distribution). If we do so, Z becomes quite sparse, indicating that some directions in the target model can be mapped to a dedicated set of features in the source model.

This observation of transferrability can potentially be extended to any pair of pre-trained models (supervised or unsupervised), preventing the need to interpret the representations of each specific model. It also gives us an important insight that various vision models, regardless of their pre-training regime, learn mostly similar concepts. To the best of our knowledge, ours is the first approach in the direction of transferring explanations across model spaces.

\section{Conclusion}
We proposed FALCON, an automatic framework to explain individual neurons in vision models. These explanations can be utilized for classification tasks (as shown in Figure \ref{fig:failure_mode}) as well as non-classification tasks like object detection and segmentation. We show that features become more interpretable when regarded in groups and propose a simple algorithm to discover all possible interpretable groups in a given representation space. 

FALCON utilizes three components: 1) A probe image dataset, 2) A large text vocabulary and 3) An off-the-shelf pre-trained vision-language encoder. With FALCON we propose a general-purpose framework, where the above components can be flexibly customized depending on the target model we wish to investigate. The concepts learned by the target model is governed by the data it was trained on. In order to explain these concepts via FALCON, we choose the probe image dataset and text vocabulary such that it is representative of the target model's training domain and encapsulates all the concepts learned by the target model. In our experiments, we use FALCON with ImageNet, LAION-400M and CLIP to explain deep models pre-trained on ImageNet-1K. These components could potentially generalize to a range of domains, since CLIP is already pre-trained on a very large scale and LAION-400M is diverse and expressive. FALCON can be updated to use even larger zero-shot vision-language encoders and vocabulary, when developed in future. To deploy FALCON on target models trained on medical images like chest x-rays, we can utilize vision-language encoders like ConVIRT \cite{zhang2020contrastive} or MedCLIP \cite{wang2022medclip}, combined with expressive vocabulary from radiology reports (ex. Mimic-cxr).

\textbf{Limitations and directions for future work:} Understanding how FALCON can be applied to explain vision-language models can be an extension of our work. Since vision-language models are trained to align representations in the vision and language space, we could potentially learn a great deal about the model's understanding by applying FALCON directly on the vision encoder. It would however be interesting to understand what information is represented uniquely by the text encoder. Understanding the equivalent of localized gradient heatmaps in the language space is still unclear and requires further research.
 
Supporting FALCON for non-image domains remains a topic for further research. Another limitation of FALCON is the requirement of a pre-trained vision-language model for the task of matching images to captions. While CLIP is trained on very diverse data and domain-specific versions of CLIP exist, there may be target models which are trained for uncommon tasks and data, that is unknown to CLIP. In our transferrability example, we show that concepts extracted from one model may be transferred to another by learning a simple linear transformation. Another important direction for future work is to test the limits of transferrability on multi-domain setups. For example, how do concepts learned by a model trained on painting images, transfer to a model trained on sketch images.

\section{Acknowledgment}
This project was supported in part by Meta grant 23010098, NSF CAREER AWARD 1942230, HR001119S0026 (GARD), ONR YIP award N00014-22-1-2271, Army Grant No. W911NF2120076, NIST 60NANB20D134, the NSF award CCF2212458, a CapitalOne grant and an Amazon Research Award. The authors would also like to thank Mazda Moayeri, Hemant Kumar, Samyadeep Basu, Sharath Gokarn, Saksham Suri, Pavan Gurudath, Nirat Saini, Pulkit Kumar and Archana Swaminathan for their help with testing our human studies. 

\bibliography{example_paper}
\bibliographystyle{icml2023}

%%%%%%%%%%%%%%%%%%%%%%%%%%%%%%%%%%%%%%%%%%%%%%%%%%%%%%%%%%%%%%%%%%%%%%%%%%%%%%%
%%%%%%%%%%%%%%%%%%%%%%%%%%%%%%%%%%%%%%%%%%%%%%%%%%%%%%%%%%%%%%%%%%%%%%%%%%%%%%%
% APPENDIX
%%%%%%%%%%%%%%%%%%%%%%%%%%%%%%%%%%%%%%%%%%%%%%%%%%%%%%%%%%%%%%%%%%%%%%%%%%%%%%%
%%%%%%%%%%%%%%%%%%%%%%%%%%%%%%%%%%%%%%%%%%%%%%%%%%%%%%%%%%%%%%%%%%%%%%%%%%%%%%%
\newpage
\appendix
% \onecolumn

\renewcommand\thefigure{\thesection.\arabic{figure}}
\setcounter{figure}{0} 

\renewcommand\thetable{\thesection.\arabic{table}}
\setcounter{table}{0} 

\section{Appendix}
We use pre-trained models from the solo-learn package \cite{solo} and the official implementation of CLIP \cite{radford2021learning}. 

\begin{table}[h]
    \centering
    \caption{\textbf{Feature groups and concepts for various models:} We tabulate the number of interpretable groups for each model and number of unique concepts extracted after explaining each group. We observe that many frequently occurring concepts are shared across models.}
    \label{tab:frequent_concepts}\
    \resizebox{0.48\textwidth}{!}{
    \begin{tabular}{c|c|c|c}
    \toprule
        \textbf{Model} & \textbf{\# interpretable groups}	& \textbf{\# unique concepts} &	\textbf{Most frequent concepts} \\
        \midrule
        \midrule
        SimSiam &	249	& 578	& 'white', 'head', 'brown', 'eye', 'face' \\
        SimCLR	& 293 & 676	& 'white', 'head', 'face', 'brown', 'blue'\\
        MoCo &	271 & 559 & 'white', 'head', 'face', 'eye', 'black' \\
        SwaV & 182 & 417 & 'head', 'brown', 'white', 'hand', 'black' \\
        BYOL & 281 & 477 & 'head', 'white', 'brown', 'eye', 'face' \\
        ResNet-50 (Sup) & 91 & 183 & 'brown', 'head', 'red', 'water', 'white' \\
        \bottomrule
    \end{tabular}
    }
\end{table}

\begin{figure}[h]
    \centering
    \includegraphics[width = 0.35\textwidth, trim = {3cm, 0cm, 0cm, 0cm}, clip]{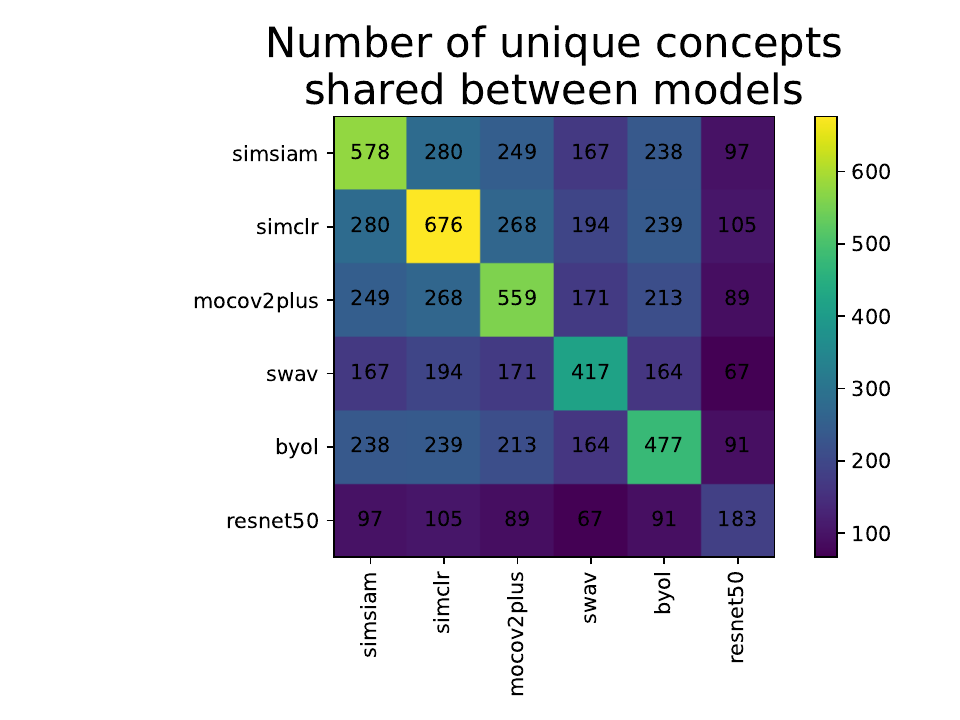}
    \caption{\textbf{Shared concepts between models:} Among the unique concepts extracted using FALCON on the representation space of various models, we plot the number shared concepts between each pair of models. }
    \label{fig:shared_concepts}
\end{figure}
\subsection{Analyzing FALCON Explanations Across Various Models}
\label{sec:global_analysis}
We have performed a global analysis comparing the FALCON concepts across various supervised and self-supervised models (ResNet-50 encoder). In Table \ref{tab:frequent_concepts}, we tabulate the number of interpretable feature groups identified from the final representation layer, along with the total number of unique concepts extracted from FALCON for these groups. Note that each explanation consists of multiple conceptual words. In the last column, we also list the most frequently occurring concepts for each model. We observe that among all the models we study, the supervised ResNet-50 model has the least number of interpretable groups and unique concepts. The most frequent concepts among all the models are almost identical, including general attributes like various colors, face, eye, which frequently occur in the ImageNet dataset. We also compute the number of shared concepts between each pair of models in Figure \ref{fig:shared_concepts}. We observe that each model shares roughly less than $50\%$ of its total concepts with any other model. This indicates that although each model is trained on the same data i.e, ImageNet, their training paradigms can enable them to encode some unique properties that are missed by other models. We calculate the number of concepts in each model that are not shared with any other model - SimSiam $160$, SimCLR $210$, MoCo $159$, SwaV $128$, BYOL $116$, ResNet-50 $39$. For example, these are some unshared (truly unique) concepts of ResNet-50 - `eel', `disc', `grip', `shooter', `tub', `sink', `weimaraner', `decal'.

\begin{table}[h]
    \centering
    \caption{\textbf{Comparing FALCON used with CLIP and LAION-400M vs BLIP-2 zero-shot captioning:} We apply FALCON with BLIP 2 \cite{li2023blip2} generated captions and ask participants to select the better explanation when compared with CLIP+LAION. BLIP captions underperform compared to CLIP+LAION. }
    \label{tab:clip_vs_blip}
    \resizebox{0.48\textwidth}{!}{
    \begin{tabular}{c|c}
    \toprule
       \textbf{Framework} & \textbf{\% of times selected as best explanation} \\
         \midrule
         \midrule
        FALCON + CLIP + LAION & \textbf{58.12} \\ 
        FALCON + BLIP 2 (OPT, caption COCO) & 41.8\\ 
         \bottomrule
    \end{tabular}
    }
\end{table}
\subsection{Employing a Captioning Model instead of CLIP}
BLIP-2's \cite{li2023blip2} zero-shot image captioning is a powerful tool to extract text captions out of highly activating images. One advantage of using a separate vocabulary with a vision-language model is the flexibility of controlling the expressiveness/specificity of the captioning dataset depending on the complexity of the target model. For example, to explain an MNIST-trained model, one may use a much smaller vocabulary whereas explaining a model like CLIP may require an equivalently large vocabulary. Moreover, the set of reference captions can be updated online, even after deployment without having to re-train any model. The similarity matrix allows us to extract multiple captions per image with a confidence score, allowing us to discard unreliable captions. Off the shelf captioning models may be domain-specific and could generate noisy captions with low expressiveness.

We compared FALCON + BLIP 2 with FALCON + CLIP + LAION in an MTurk evaluation over $91$ features and asked participants to select the best describing explanation for the displayed set of images (See Table \ref{tab:clip_vs_blip}. Explanations generated via our CLIP+LAION captioning outperforms BLIPs captioning, however, BLIP 2 is still a practical alternative given that it is trained on a large scale on LAION. 

\begin{table*}[h]
    \centering
    \caption{\textbf{Percentage of highly activating features in the ResNet-50 representation space:} For different model representations, we tabulate the percentage of features that activate at least 10 samples with a magnitude greater than $\alpha$. We select $\alpha$ according to the mean of the distribution of the representation space (See Section \ref{sec:which_features} for more details). }
    \resizebox{\textwidth}{!}{
    \begin{tabular}{c|c|c|c|c|c|c|c}
    \toprule
         \textbf{Model} & \textbf{ResNet-50 (Supervised)} & \textbf{SimCLR} & \textbf{MoCo} & \textbf{DINO} & \textbf{BYOL} & \textbf{SimSiam} & \textbf{SwaV} \\
         \midrule
         \midrule
         \textbf{\% features that highly activate $>$ 10 samples} & 0.68 & 21.92 & 17.70 & 16.66 & 25.63 & 8.00 & 6.00 \\
         $\mathbf{\alpha}$ & 0.27 & 0.34 & 0.34 & 0.14 & 0.24 & 0.32 & 0.31 \\
    \bottomrule
    \end{tabular}
    }
    \label{tab:count-features}
\end{table*}
\begin{figure}
    \centering
    \subfigure{\includegraphics[width=0.23\textwidth]{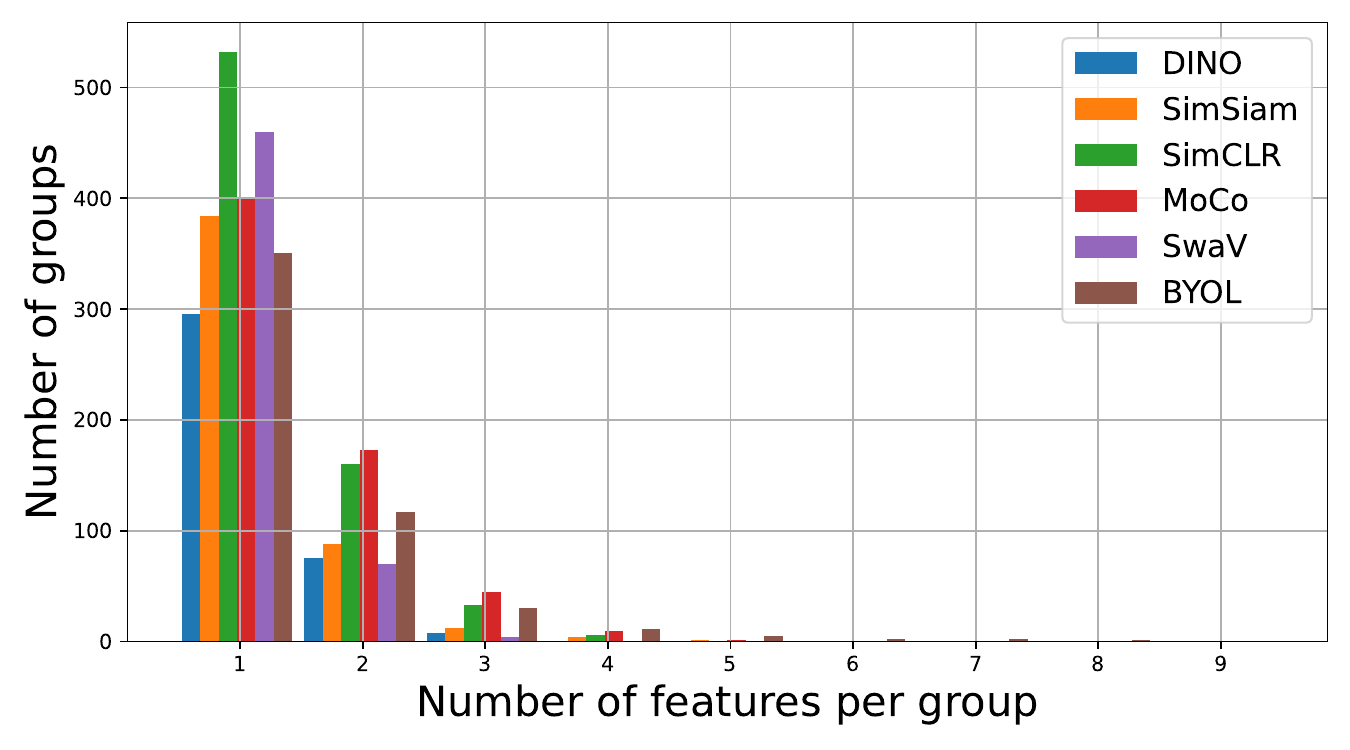}}
    \subfigure{\includegraphics[width=0.23\textwidth]{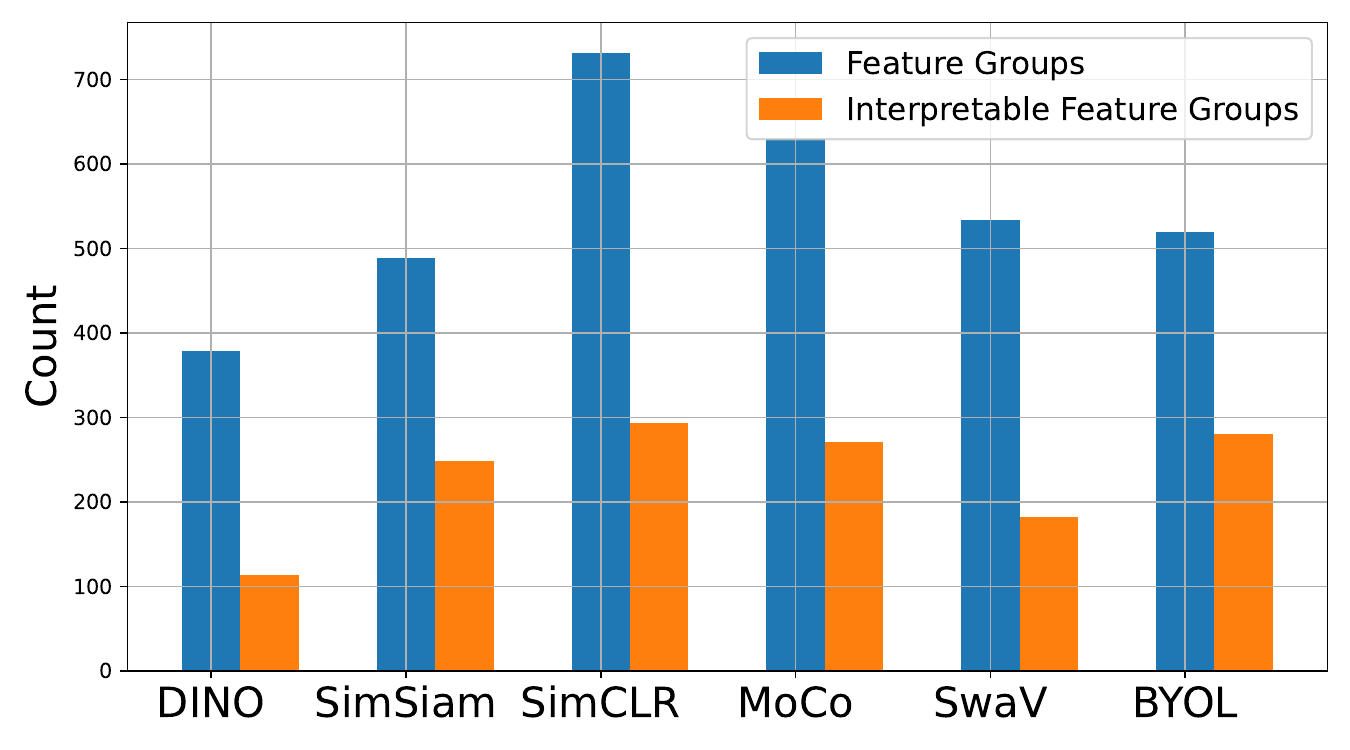}}
    \caption{\textbf{Distribution of feature groups:} For different self-supervised model representation spaces we compute the feature groups (from Algorithm \ref{alg:int_feat_selection}. On the left we plot group sizes against the number of groups and on the right, we plot the number of interpretable groups among the discovered feature groups.} 
    \label{fig:feat_groups}
\end{figure}
\subsection{Interpretable Features in Various Models}
\begin{figure}[h]
    \centering
    \includegraphics[width = 0.48\textwidth]{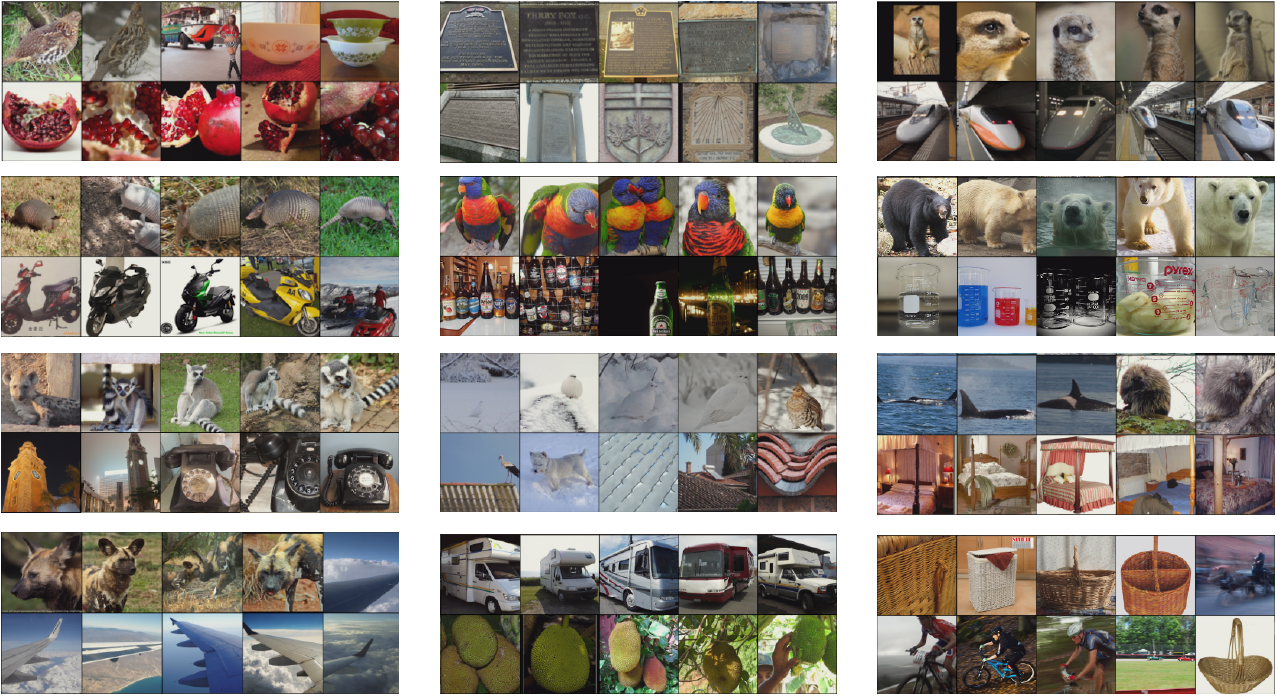}
    \caption{\textbf{Examples of top activating images of some un-explainable independent features:} We provide more examples of top activating images of some independent features of DINO \cite{dino} (ResNet-50 \cite{resnet}) representations. The image sets are not correlated in any sense, making it hard to discover shared concepts for these features.}
    \label{fig:non_explainable}
\end{figure}

We discuss in Section \ref{sec:which_features} that to discover potentially explainable features we can apply a strong value for $\alpha$ in $\mathcal{T}_i$, set of highly activating images. Since the distribution of each model representation space can be different, to be consistent we select $\alpha = mean(\bH) + 16 \times std(\bH)$ (where $\bH$ is the representation matrix). In Figure \ref{tab:count-features}, we tabulate the percentage of highly activating features in the final-layer representations where $|\mathcal{T}_i| > 10$. ResNet-50 has a particularly low number of highly activating features compared to self-supervised baselines. The remaining features in the representation space (or by relaxing $\alpha$ to be less rigorous), do not activate a resembling set of images, making such features harder to explain (Figure \ref{fig:comb_features}). Some more examples of such features are shown in Figure \ref{fig:non_explainable}.

We also discussed in Section \ref{sec:which_features} that simply thresholding by $\alpha$ does not guarantee explainability as the top activating images can still be unrelated. A larger portion of the representation space can be explained with feature groups. Using the Algorithm \ref{alg:int_feat_selection}, we discover feature groups and interpretable feature groups for various self-supervised models. In Figure \ref{fig:feat_groups}, on the left, we show the distribution of feature groups and their size. All the identified groups contain at least $10$ highly activating images. A large percentage of feature groups contain 1-2 features per group, however, there also exist feature groups that contain up to 9 features. On the right, we compare the feature groups and the \textit{interpretable} feature groups, according to Algorithm \ref{alg:int_feat_selection}. The interpretable groups activate samples that are more similar (based on CLIP cosine similarity) and are therefore easy to explain with shared natural language concepts.

\subsection{Human Study to Evaluate Concepts}
\label{sec:mturkstudy}
\textbf{Eliminating malicious and inadequate responses:} In our studies, we only select participants that have a HIT approval rate of greater than $90$ and the number of HIT approvals is $> 500$ in the past. Each task is active for $30$ minutes allowing the participants ample time to make their selections. We did not explicitly include control questions, however, we identified a small number of tasks which had low-quality concept sets which we used to verify the reliability of the participants. We also approve and pay the participants only after verifying their annotation quality.

\begin{figure}[h]
    \centering
    \includegraphics[width = 0.4\textwidth]{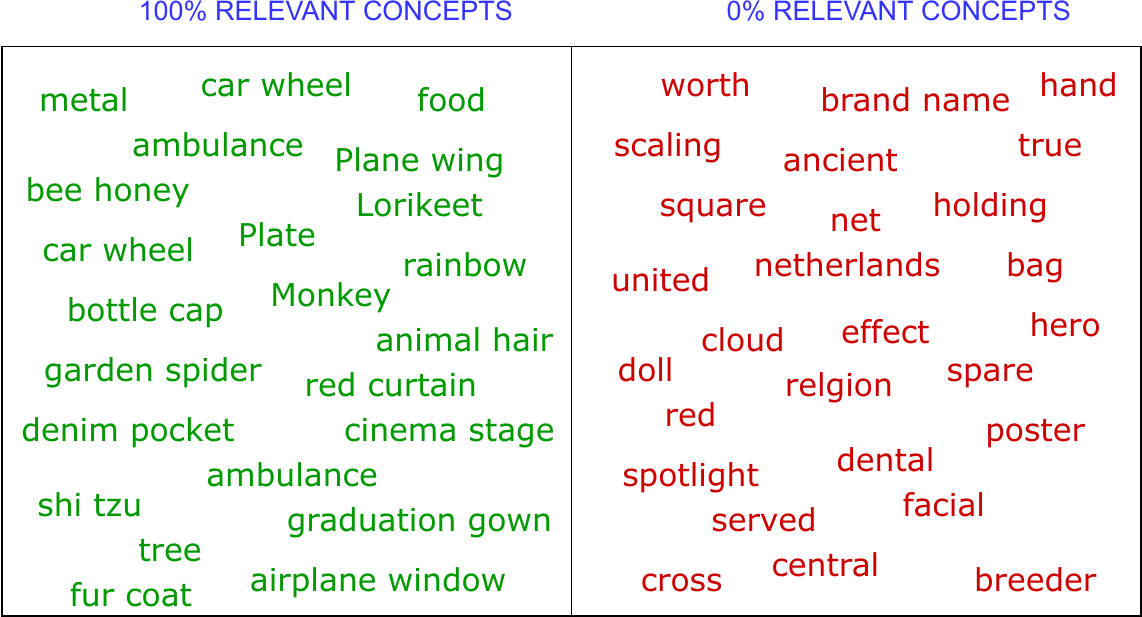}
    \caption{\textbf{Comparing most and least relevant concepts based on AMT study:} We display the concepts with $100\%$ relevancy agreement on the left and the concepts with $0\%$ relevancy agreement on the right. }
    \label{fig:relevant_vs_non_relevant}
\end{figure}
\begin{figure*}[h]
    \centering
    \includegraphics[width = 0.8\textwidth]{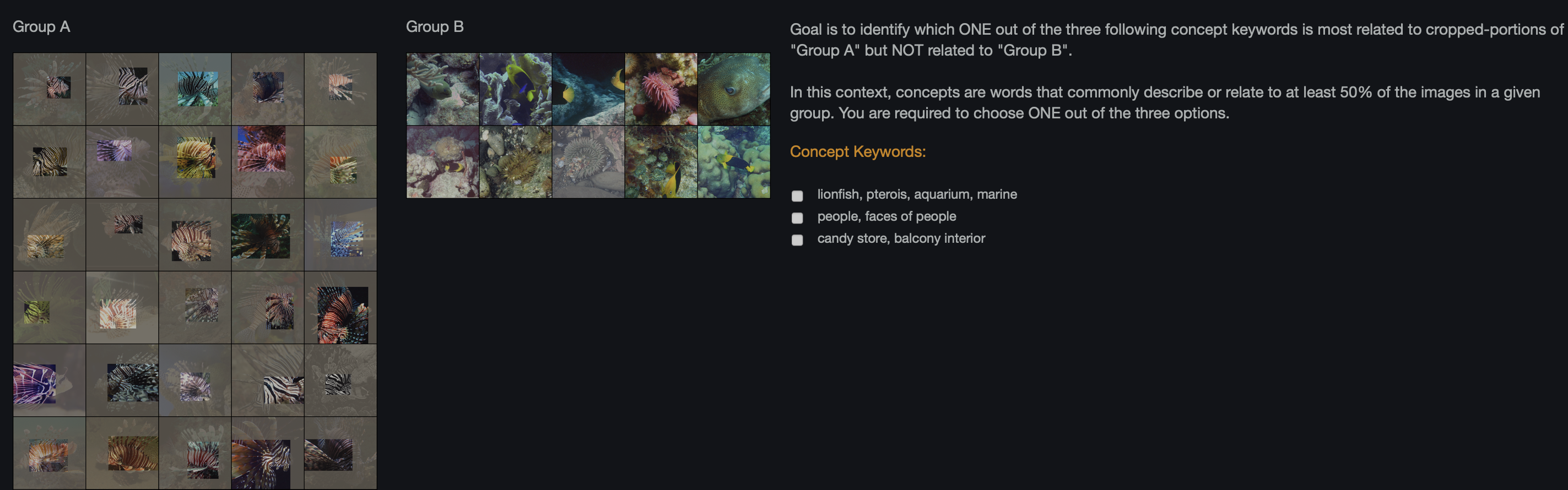}
    \caption{\textbf{Amazon Mechanical Turk user study template:} A template of our user study where we display two groups of images for a target feature and ask the users to select the best explanation among $3$ options.}
    \label{fig:mturk}
\end{figure*}
\begin{figure}[h]
    \centering
    \includegraphics[width = 0.45\textwidth]{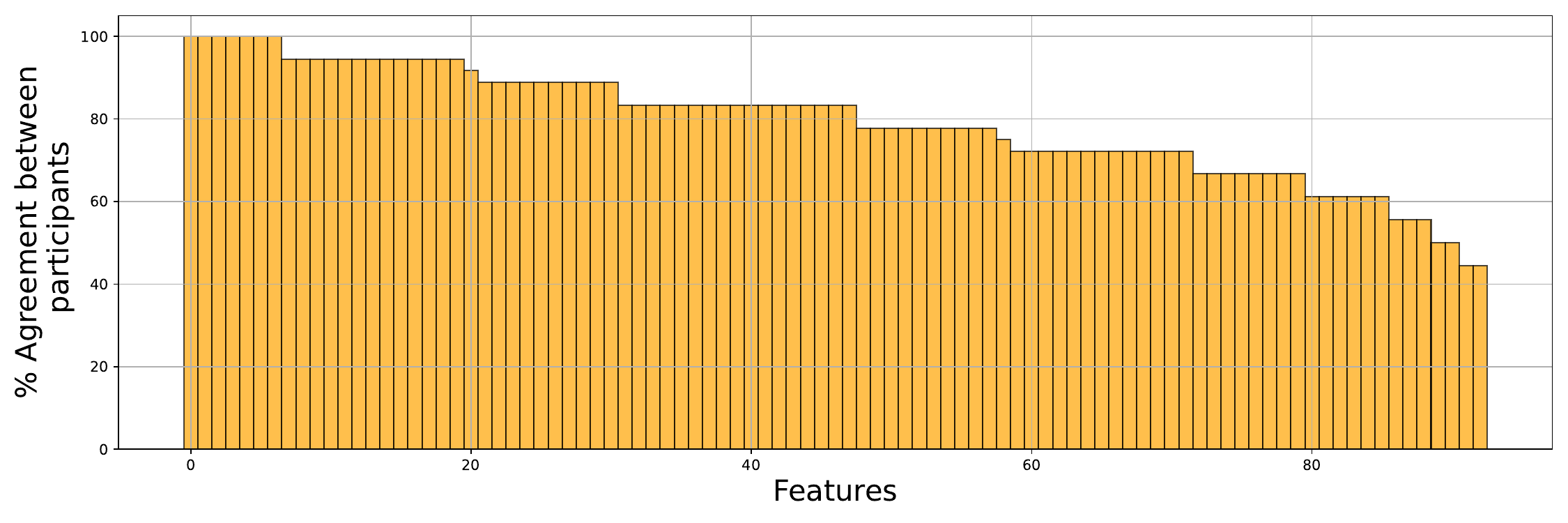}
    \caption{\textbf{Average agreement between participants for each feature:} We plot the agreement of relevancy for each concept averaged by the feature for $93$ features we perform human study on. }
    \label{fig:agreement}
\end{figure}
In Figure \ref{fig:mturk}, we show a template of our user study where we display two groups i.e., highly activated cropped images (Group A) and lowly activating images (Groups B). In this example, we compare FALCON concepts to that of MILAN and Net-Dissect. As discussed in Section \ref{sec:evaluation}, we also evaluate top $6$ FALCON concepts on their relevancy. We define the \textit{agreement of relevancy} between workers as the percentage of workers that believe a concept is relevant. This, averaged for all concepts in a feature, is plotted in Figure \ref{fig:agreement}. We observe that, for $93$ features, up to $86\%$ of them are agreed to be relevant among at least $66 \%$ of the workers. We also visualize the concepts where the agreement of relevancy is $100\%$ (left) and $0\%$ (right) in Figure \ref{fig:relevant_vs_non_relevant}. We observed that the irrelevant concepts have a very low average CLIP score of $0.067$. This is likely because there were other, more specific concepts for that feature, or the concepts were out-of-context for the displayed images. In contrast, the concepts with $100\%$ relevancy have a relatively higher average CLIP score of $0.284$ (unsurprisingly) and are strongly correlated with the displayed images.

\subsection{Transferring Concepts to Unseen Data}
\begin{figure*}[h]
    \centering
    \includegraphics[width = 0.9 \textwidth]{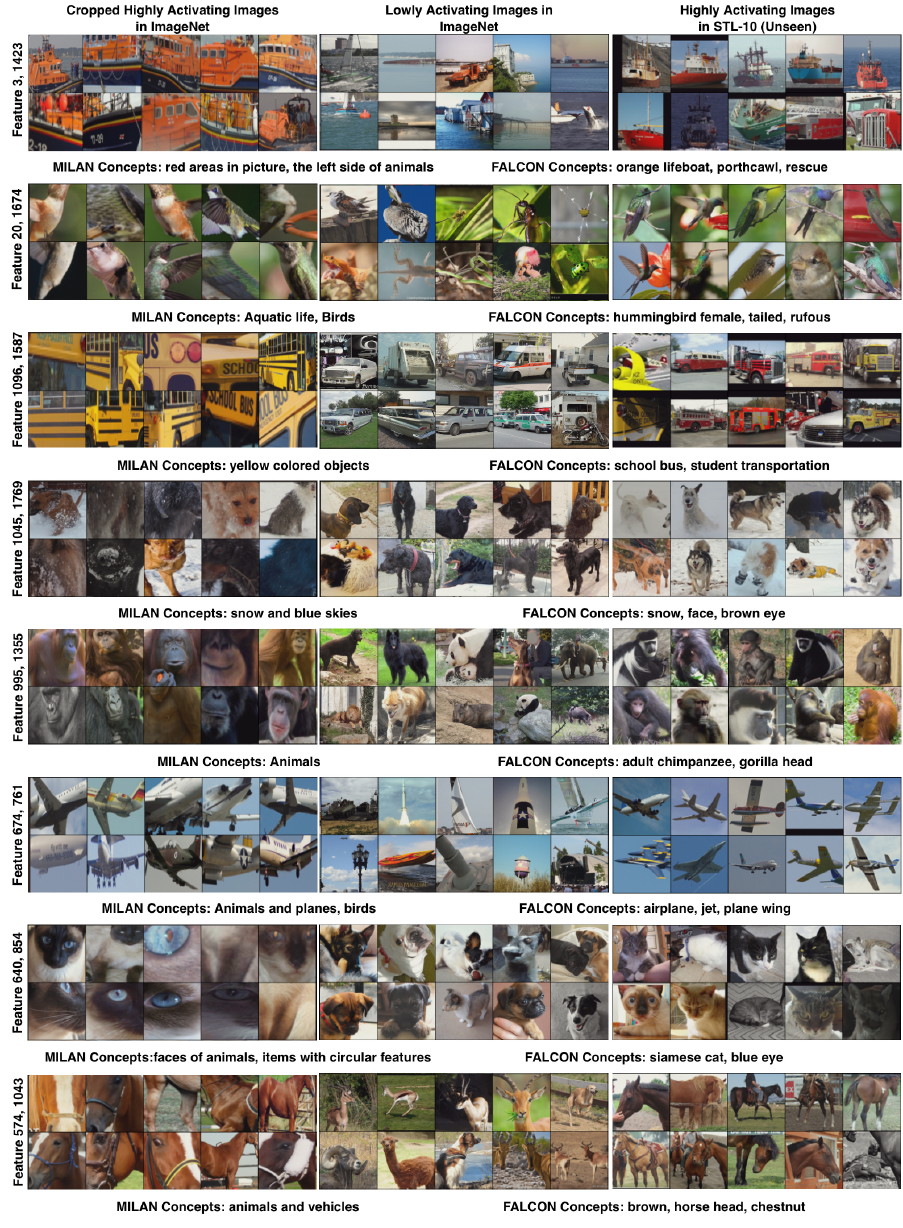}
    \caption{\textbf{Generalization of concepts to unseen data:} We extract concepts from various features of DINO \cite{dino} representations (using ImageNet) and verify if they generalize to STL-10 \cite{stl10}, an unseen dataset. In all features, the STL-10 images closely resemble the ImageNet images and contain the concepts described by FALCON.}
    \label{fig:concepts_unseen_data}
\end{figure*}

In Section \ref{sec:transfer}, we study a non-trivial setup of transferring concepts from one interpretable model to another. In this Section we study a simpler scenario of transferring concepts to unseen datasets. Essentially, we evaluate if our extracted concepts (on ImageNet validation data), generalizes to new datasets. In Figure \ref{fig:concepts_unseen_data}, for several DINO features, we display the highly (cropped) and lowly activating images, as well as the highly activating images in STL-10 \cite{stl10} which is an unseen dataset. We extract concepts using FALCON and MILAN to compare the quality. We observe that STL-10 images for each feature closely resemble that of ImageNet and more importantly, correlate with most of FALCON concepts. FALCON also generally provides more explicit concepts covering multiple aspects, compared to MILAN. This confirms that extracted concepts generalize well to unseen or unknown data.

\subsection{Explaining Supervised Representations and Early-Layer Features}
\begin{figure*}
    \centering
    \includegraphics[width = 0.9\textwidth]{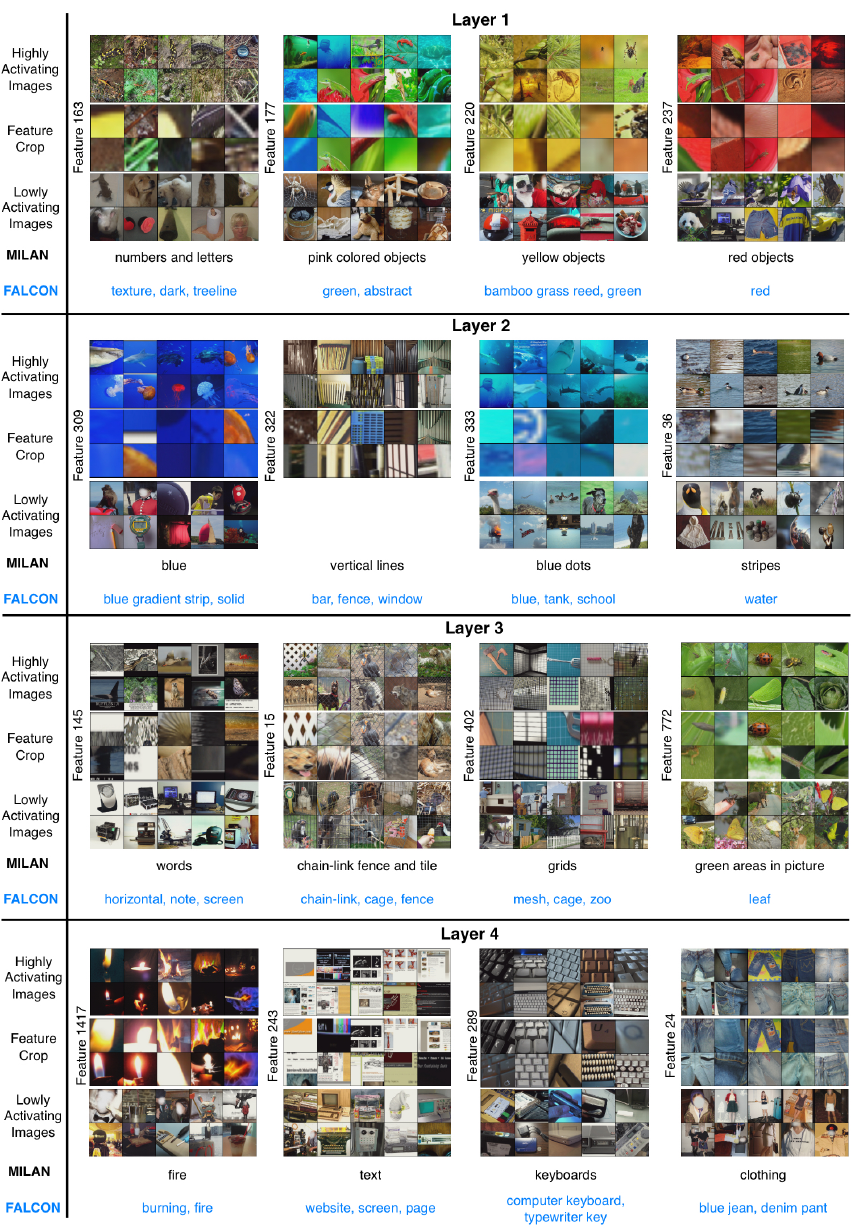}
    \caption{\textbf{Concepts for features of various layers of supervised ResNet-50:} We extract concepts from random features of layers of supervised pre-trained ResNet-50. We compare FALCON concepts with MILAN concepts. }
    \label{fig:resnet50}
\end{figure*}

To further confirm the generalizability of our concept extraction framework, FALCON, we extract concepts from different layers of supervised pre-trained ResNet-50 (using ImageNet and LAION). Our results are shown in Figure \ref{fig:resnet50}. We observe the initial layer features, activate very primitive type concepts like color or geometric patters. FALCON extracts this information in its concepts based on the cropped images. As we move closer to the final layer, the feature crops become larger and concepts become more descriptive. We thus confirm that FALCON can be applied to explain any neuron in any vision model, supervised or unsupervised.

%%%%%%%%%%%%%%%%%%%%%%%%%%%%%%%%%%%%%%%%%%%%%%%%%%%%%%%%%%%%%%%%%%%%%%%%%%%%%%%
%%%%%%%%%%%%%%%%%%%%%%%%%%%%%%%%%%%%%%%%%%%%%%%%%%%%%%%%%%%%%%%%%%%%%%%%%%%%%%%

\end{document}